\documentclass[10pt,letterpaper,twocolumn]{article}
\usepackage[latin9]{inputenc}
\usepackage{color}
\usepackage{array}
\usepackage{multirow}
\usepackage{amsmath}
\usepackage{amssymb}
\usepackage{graphicx}
\usepackage[unicode=true,
 bookmarks=false,
 breaklinks=true,pdfborder={0 0 1},backref=section,colorlinks=false]
 {hyperref}
\hypersetup{
 pagebackref=true,letterpaper=true,colorlinks}

\makeatletter

\providecommand{\tabularnewline}{\\}

\usepackage{iccv}
\usepackage{times}
\usepackage{epsfig}
\usepackage{graphicx}

\iccvfinalcopy 

\def\iccvPaperID{2650} 

\ificcvfinal\fi

\makeatother

\begin{document}

\title{A Fine-Grained Facial Expression Database for End-to-End Multi-Pose Facial Expression Recognition}

\author{Wenxuan Wang\\
Fudan University\\
{\tt\small wxwang17@fudan.edu.cn}
\and
Qiang Sun\\
Fudan University\\
{\tt\small 18110860051@fudan.edu.cn}
\and
Tao Chen\\
Fudan University\\
{\tt\small eetchen@fudan.edu.cn}
\and
Chenjie Cao\\
Ping An OneConnect\\
{\tt\small caochenjie948@pingan.com}
\and
Ziqi Zheng\\
Ping An OneConnect\\
{\tt\small zhengziqi356@pingan.com.cn}
\and
Guoqiang Xu\\
Ping An OneConnect\\
{\tt\small xuguoqiang371@pingan.com.cn}
\and
Han Qiu\\
Ping An OneConnect\\
{\tt\small hannaqiu@pingan.com.cn}
\and
Yanwei Fu *\\
Fudan University\\
Corresponding Author\\
{\tt\small yanweifu@fudan.edu.cn}
\footnote{\noindent \textbf{*}: Corresponding Author}
}

\maketitle
\begin{abstract}
The recent research of facial expression recognition has made a lot of progress due to the development of deep learning technologies, but some typical challenging problems
such as the variety of rich facial expressions and poses are still not resolved. To
solve these problems, we develop a new Facial Expression Recognition
(FER) framework by involving the facial poses into our image synthesizing
and classification process. There are two major novelties in this
work. First, we create a new facial expression dataset of more than
200k images with 119 persons, 4 poses and 54 expressions. To our knowledge
this is the first dataset to label
faces with subtle emotion changes for expression recognition purpose. It is also the first dataset
that is large enough to validate the FER task on unbalanced poses, expressions, and zero-shot subject IDs.
Second, we propose a facial pose generative adversarial network
(FaPE-GAN) to synthesize new facial expression images to augment the
data set for training purpose, and then learn a LightCNN based
Fa-Net model for expression classification. Finally, we advocate four novel
learning tasks on this dataset. The experimental results well validate the effectiveness of the proposed
approach.
\end{abstract}

\section{Introduction}

Facial expression \cite{corneanu2016survey}, as the most
important facial attribute, reflects the emotion status of a person,
and contains meaningful communication information. Facial expression
recognition (FER) is widely used in multiple applications such as
psychology, medicine, security and education \cite{corneanu2016survey}.
In psychology, it can be used for depression recognition for analyzing
psychological distress. On the other hand, detecting a student's concentration
or frustration is also helpful in improving the educational approach.

Facial expression recognition mainly contains four steps: face detection,
face alignment, feature extraction and facial expression classification.
(1) In the first step, the face is detected from the image with each
labelled by a bounding box. (2) In the second step, the face landmarks
are generated to align the face. (3) In the third step, the features
that contain facial related information are extracted in either hand-crafted
way, \emph{e.g.}, SIFT, \cite{berretti20113d} Gabor wavelets
\cite{bartlett2005recognizing,lyons1998coding} and LBP \cite{shan2009facial}
or learned way by a neural network. (4) In the fourth step,
various classifiers such as SVM, KNN and MLP can be adopted for facial expression classification.

The recent renaissance of deep neural networks delivers the human
level performance towards several vision tasks, such as object classification,
detection and segmentation \cite{liu2015deep,liu2017hydraplus,qian2017multi}. Inspired by this, some deep
network methods \cite{khorrami2015deep,minaee2019deep,zhang2018joint}
have been proposed to address the facial expression recognition.
In FER task, facial expression is usually assumed to
contain six discrete primary emotions: anger, disgust, fear, happy,
sad and surprise according to Ekman's theory. With an additional neutral
emotion, the seven emotions compose the main part of most common
emotion datasets, including CK+ \cite{lucey2010extended,kanade2000comprehensive},
JAFEE \cite{lyons1998coding}, FER2013 \cite{fer2013} and FERG \cite{aneja2016modeling}.

However, one most challenging problem of FER in fact is lacking of a large-scale dataset of high quality images, that can
be employed to train the deep networks and
investigate the impacting factors for the FER task. Another
disadvantage of these datasets, \emph{e.g.}, JAFFE and FER2013 dataset,
is the little diversity of expression emotions,
which cannot really express the  versatile  facial expression emotions in the real world life.

To this end, we create a new dataset $\mathrm{F}^{2}$ED (Fine-grained
Facial Expression Database) with 54 emotion types, which include larger
number of emotions with subtle changes, such as calm, embarrassed,
pride, tension and so on. Further, we also consider the influence
of face pose changes on the expression recognition, and introduce
the pose as another attribute for each expression. Four orientations
(poses) including front, half left, half right and bird view are labelled,
and each has a balanced number of examples to avoid training bias.

On this dataset, we can further investigate how the poses, expressions,
and subject IDs affect the FER performance. Critically, we propose
four novel learning tasks over this dataset as shown in Fig.~\ref{fig:medb}(c).
They are expression recognition with the standard balanced setting
(ER-SS), unbalanced expression (ER-UE), unbalanced poses (ER-UP),
and zero-shot ID (ER-ZID). Similar to
the typical zero-shot learning setting \cite{lampert2014attribute}, the zero-shot ID setting means that the testing faces of persons have not appeared in the training set.
To tackle these four learning tasks, we further design a novel framework
that can augment training data, and then train the classification
network. Extensive experiments on our dataset, as well as JAFEE \cite{lyons1998coding},
FER2013 \cite{fer2013} show that (1) our dataset is large enough
to be used to pre-train a deep network as the backbone network; (2) the unbalanced poses,
expressions and zero-shot IDs indeed negatively affect the FER
task; (3) the data augmentation strategy is helpful to learn a more powerful model yielding better performance. These three points are also the main
contributions of this paper.

\section{Related Work}

\subsection{Facial expression recognition}

Extensive FER works based on neural networks have been proposed \cite{khorrami2015deep,wang2017multi,zhang2016joint}.
Khorrami \emph{et al.} \cite{khorrami2015deep} trains a CNN for FER
task, visualizes the learned features and finds that these features
strongly correspond to the FAUs proposed in \cite{ekman1997face}.
Attentional CNN \cite{minaee2019deep} on FER is proposed to focus
on the most salient parts of faces by adding a spatial transformer.

Generative Adversarial Net (GAN) \cite{goodfellow2014generative}
based models have also been investigated in solving the FER task.
Particularly, GAN is usually composed of a generator and a discriminator. In order to weaken the influence of pose and occlusion,
the pose-invariant model \cite{zhang2018joint} is proposed by generating
different pose and expression faces based on GAN. Qian \emph{et al.} \cite{qian2018pose} propose a generative adversarial network (GAN) designed specifically for pose normalization in re-id.
Yan \emph{et al.}
\cite{yang2018facial} propose a de-expression model to generate
neutral expression images from source images by Conditional cGAN \cite{mirza2014conditional}, and use the
residual information in the intermediate layer in GAN to classify
the expression.

\begin{figure*}
\centering{}%
\begin{tabular}{ccc}
\hspace{-0.2in}%
\begin{tabular}{c}
\includegraphics[width=0.45\columnwidth]{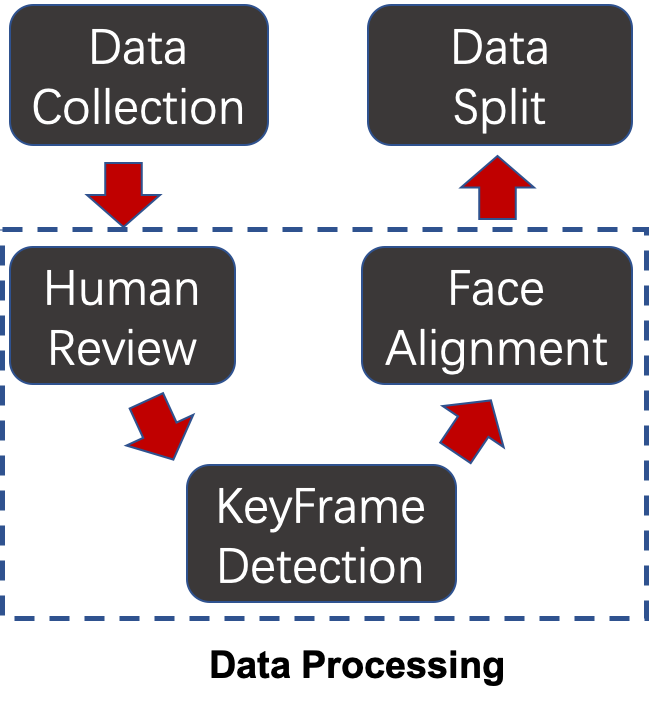}\tabularnewline
\end{tabular} & \hspace{-0.2in}%
\begin{tabular}{c}
\includegraphics[width=0.35\textwidth]{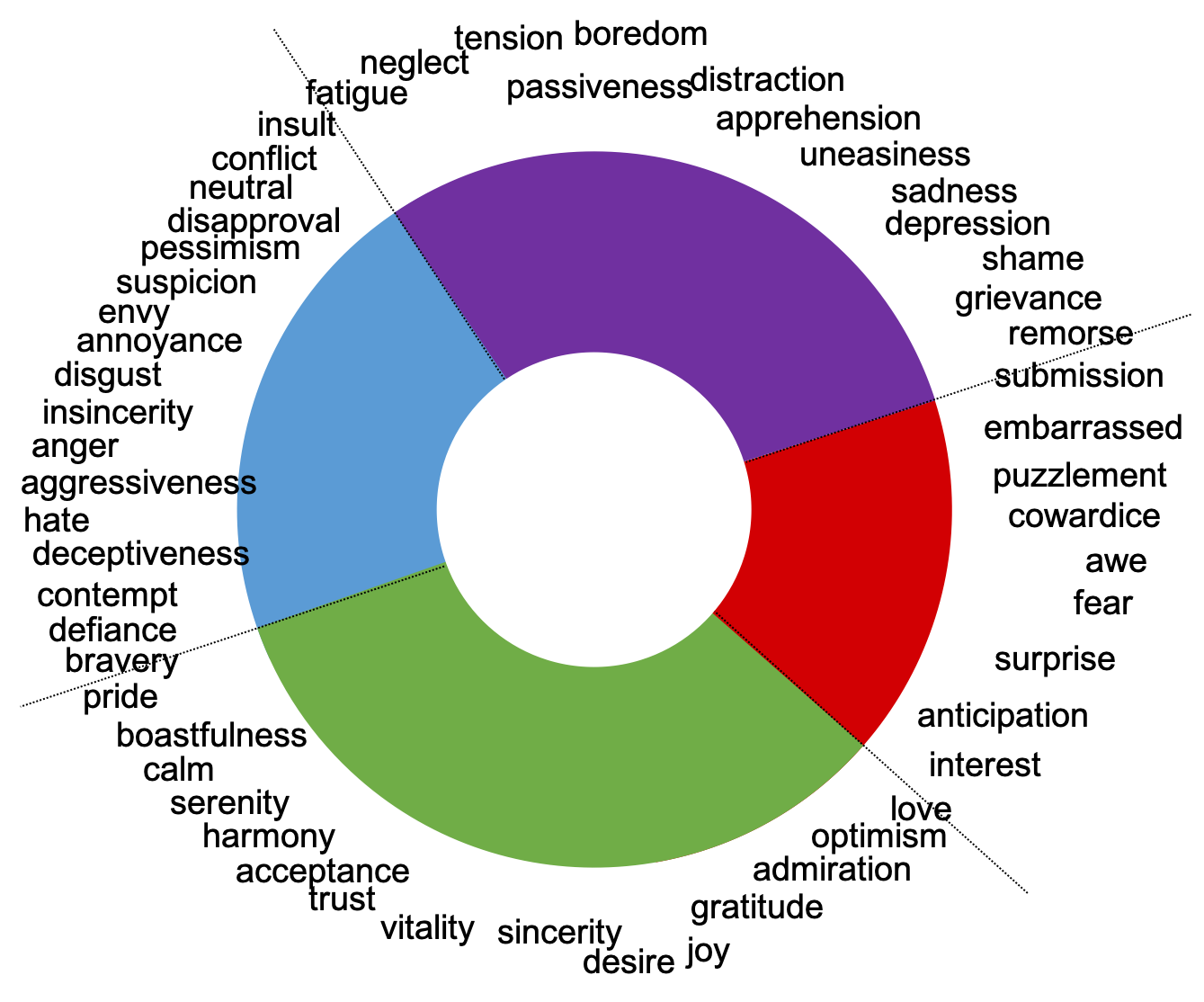}\tabularnewline
\end{tabular} & \hspace{-0.2in}%
\begin{tabular}{c}
\includegraphics[width=0.3\textwidth]{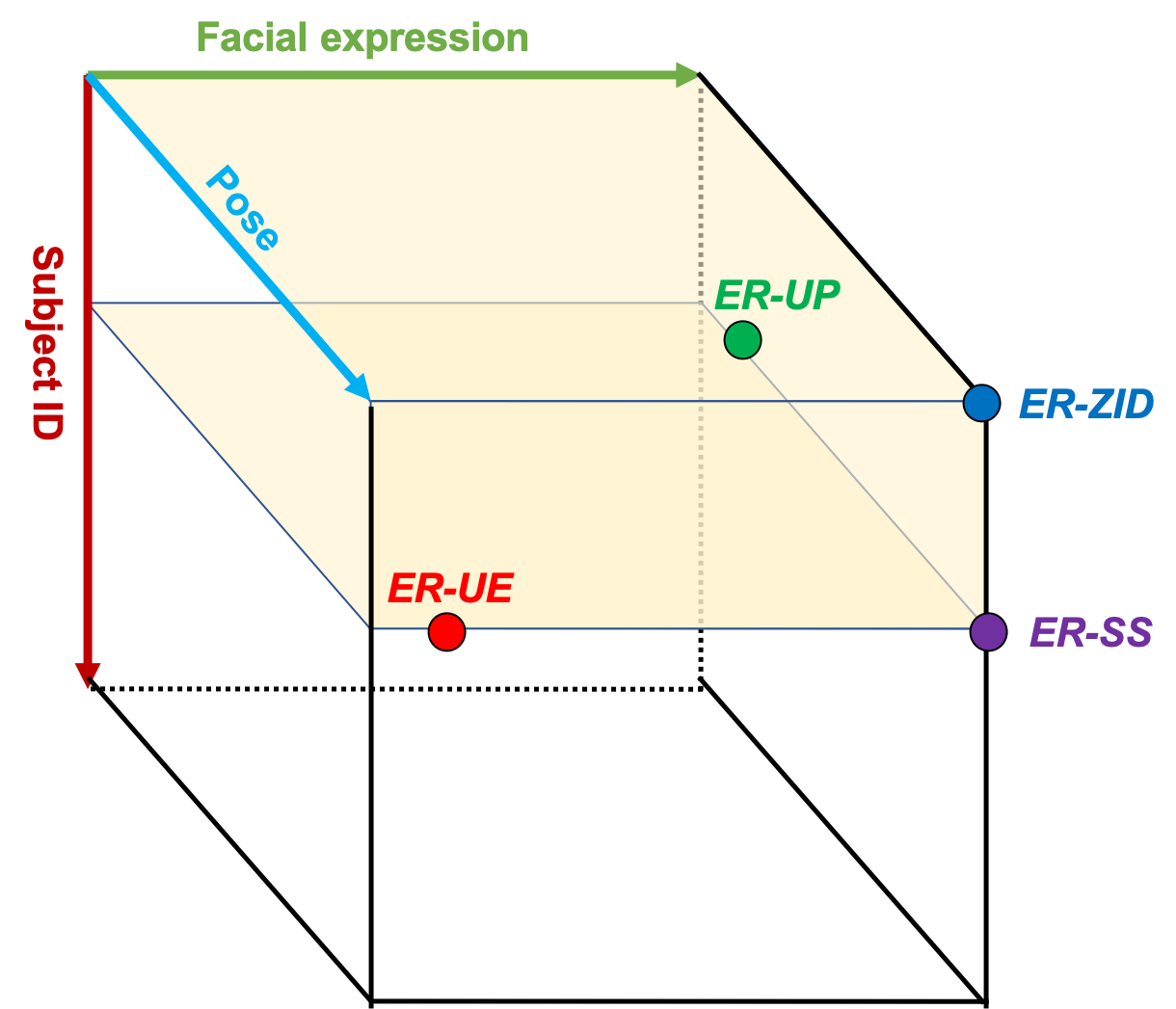}\tabularnewline
\end{tabular}\tabularnewline
(a) Data processing  & (b) Expression classes & (c) Problem Context \tabularnewline
\end{tabular}\caption{\label{fig:medb} (a) We show the flow of data processing of $\mathrm{F}^{2}$ED
dataset. (b) $\mathrm{F}^{2}$ED has 54 different facial expression
classes, and we organize them into four large classes. (c) $\mathrm{F}^{2}$ED
dataset can be applied to various problem contexts. ER-SS: Expression recognition in the standard setting, ER-UE: Expression recognition with unbalanced expression, ER-UP: Expression recognition with unbalanced poses, ER-ZID: Expression recognition with zero-shot ID.}
\end{figure*}

\subsection{Previous Datasets}

\noindent \textbf{CK+}. The extended Cohn-Kanade (CK+) database \cite{lucey2010extended}is
an updated version of CK database \cite{kanade2000comprehensive}.
In CK+ database, there are 593 video sequences from 123 subjects.
Of the 593 video sequences, 327 are selected according to the FACS
coded emotion labels. The last frame of the selected video is labeled
as one of the eight emotions: angry, contempt, disgust, fear, happy,
sad, surprise and neutral.

\noindent \textbf{JAFFE}. The Japanese Female Facial Expression (JAFFE)
database \cite{lyons1998coding} contains 213 images of 256$\times$256
pixels resolution. The images are taken from 10 Japanese female models
in a controlled environment. Each image is rated with one of the following
6 emotion adjectives: angry, disgust, fear, happy, sad and surprise.

\noindent \textbf{FER2013}. The Facial Expression Recognition 2013
database \cite{fer2013} contains 35887 images of 48$\times$48 resolution.
These images are taken in the wild setting which means more challenging
conditions such as occlusion and pose variations are included. They
are labelled as one of the seven emotions as described above. The dataset
is split into 28709 training images, 3589 validation images and 3589
test images.

\noindent \textbf{KDEF}. The dataset of Karolinska Directed Emotional Faces
\cite{lundqvist1998karolinska} contains 4900 images of $562\times762$
pixels resolution. The images are taken from 140 persons (70 male,
70 female) from 5 angles with 7 emotions. The angles contain full
left profile, half left profile, front, full right profile and half right profile. The emotion set contains 7 expressions: afraid, angry,
disgusted, happy, sad, surprised and neutral.

\subsection{Learning paradigms }

Zero-shot learning recognize the new visual categories that have not
been seen in the labelled training examples \cite{lampert2014attribute}.
The problem is usually solved by transferring learning from source
domain to the target domain. Semantic attributes that describe a new
object can be utilized in zero-shot learning. Xu \emph{et al}.\cite{xu2018heterogeneous}
propose a zero-shot video emotion recognition. In this paper, we
propose a novel FER task on the persons that are not in the training
set.  On the other hand, class imbalance is a common
problem, especially in deep learning \cite{inbalanced_data,giannopoulos2018deep}.
For the first time, we propose a dataset that is large enough to help to
evaluate the influence of unbalanced poses, expressions, and person
IDs over the FER task. To alleviate this issue, we investigate synthesizing
more data by GAN-based data augmentation inspired by recent works
on Person Re-ID\cite{qian2018pose} and Facial expression recognition
\cite{zhang2018joint}.

\section{Fine-Grained Facial Expression Database}

\begin{table*}
\centering{}{\small{}}%
\begin{tabular}{c|c|c|c|c|c|c|c|c}
\hline
{\small{}dataset} & {\small{}\#expression} & {\small{}\#subject} & {\small{}\#pose} & {\small{}\#image} & {\small{}\#sequence} & {\small{}Resolution} & {\small{}Pose list} & {\small{}Condition}\tabularnewline
\hline
\hline
{\small{}CK+} & {\small{}8} & {\small{}123} & {\small{}1} & {\small{}327} & {\small{}593} & {\small{}$490\times640$} & {\small{}F} & {\small{}Controlled}\tabularnewline
\hline
{\small{}JAFFE} & {\small{}7} & {\small{}10} & {\small{}1} & {\small{}213} & {\small{}-} & {\small{}$256\times256$} & {\small{}F} & {\small{}Controlled}\tabularnewline
\hline
{\small{}FER2013} & {\small{}7} & {\small{}-} & {\small{}-} & {\small{}35887} & {\small{}-} & {\small{}$48\times48$} & {\small{}-} & {\small{}In-the-wild}\tabularnewline
\hline
{\small{}KDEF} & {\small{}7} & {\small{}140} & {\small{}5} & {\small{}4900} & {\small{}-} & {\small{}$562\times762$} & {\small{}FL,HL,F,FR,HR} & {\small{}Controlled}\tabularnewline
\hline
\hline
{\small{}$\mathrm{F}^{2}$ED} & {\small{}54} & {\small{}119} & {\small{}4} & {\small{}219719} & {\small{}5418} & {\small{}$256\times256$} & {\small{}HL,F,HR,BV} & {\small{}Controlled}\tabularnewline
\hline
\end{tabular}\caption{\label{tab:Comparison-with-existing}Comparison $\mathrm{F}^{2}$ED
with existing facial expression database. In the pose list, F : front,
FL : full left, HL: half left, FR: full right, HR: half right, BV:
bird view }
\end{table*}
To the best of our knowledge, we contribute the largest fine-grained
facial expression dataset to the community. Specifically, our $\mathrm{F}^{2}$ED
dataset has the largest number of images (totally 219719 images) with
119 identities and 54 kinds of fine-grained facial emotions. Each
person is captured from four different views of cameras as shown in
Fig.~\ref{fig:Cameras-used-to}. Furthermore, in Tab.~\ref{tab:Comparison-with-existing},
our dataset is compared against the existing dataset -- CK+, JAFFE,
FER2013, KDEF. We show that our $\mathrm{F}^{2}$ED is orders of magnitude
larger than these existing datasets in terms of expression classes and number of total images.

\begin{figure*}
\centering{}\includegraphics[width=0.8\textwidth]{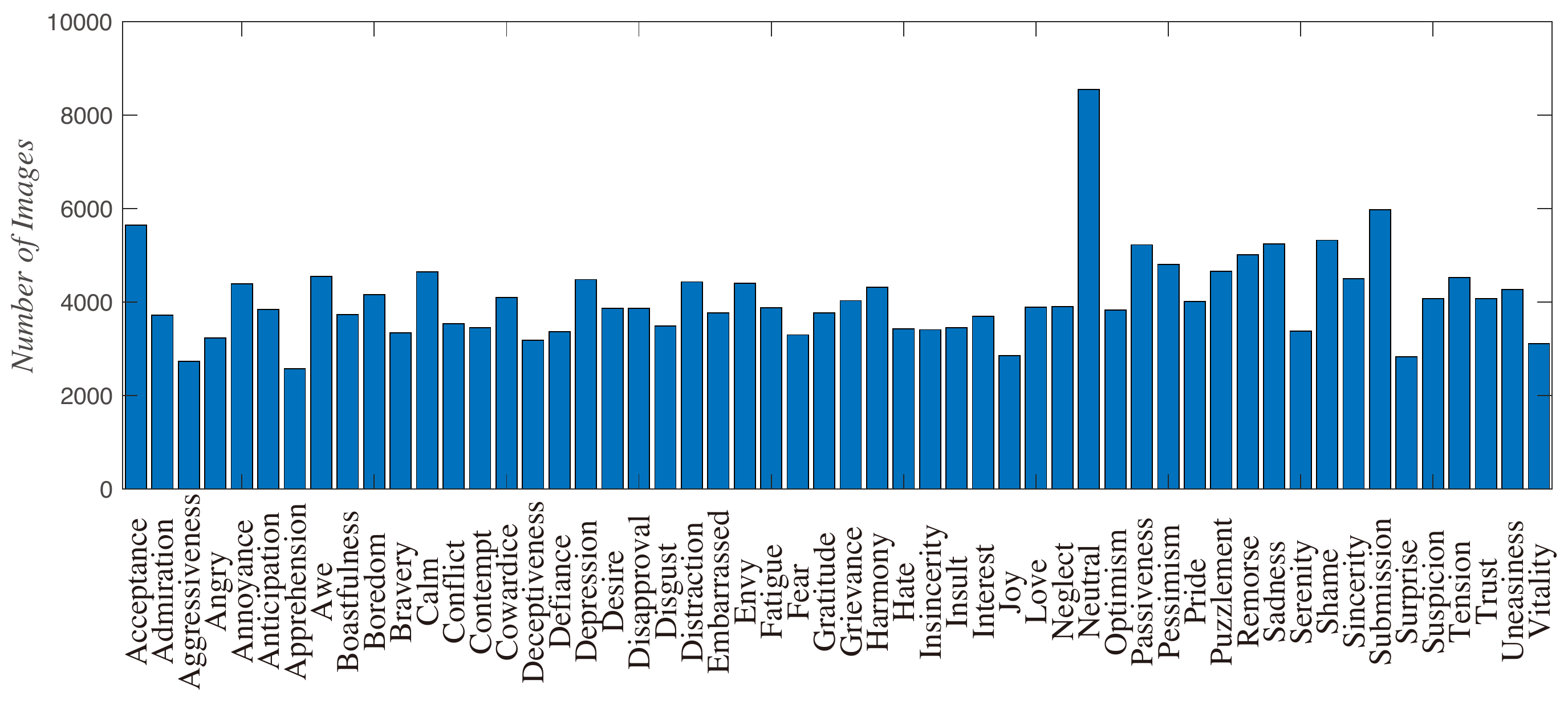}\caption{\label{fig:dist_on_expr}Image distribution of different expressions.}
\end{figure*}

\subsection{The collection of $\mathrm{F}^{2}$ED }

\begin{figure}
\begin{centering}
\begin{tabular}{cc}
\hspace{-0.3in}%
\begin{tabular}{c}
\includegraphics[width=0.15\textwidth]{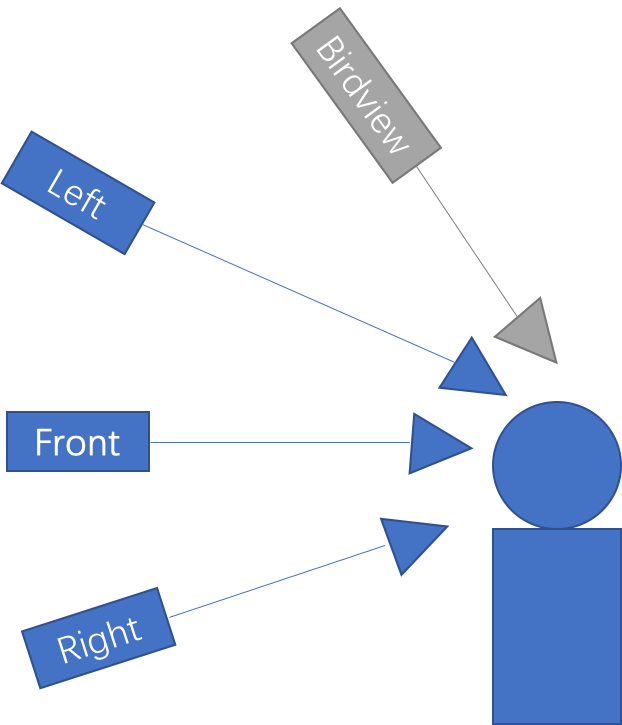}\tabularnewline
\end{tabular} & \hspace{-0.3in}%
\begin{tabular}{c}
\includegraphics[width=0.35\textwidth]{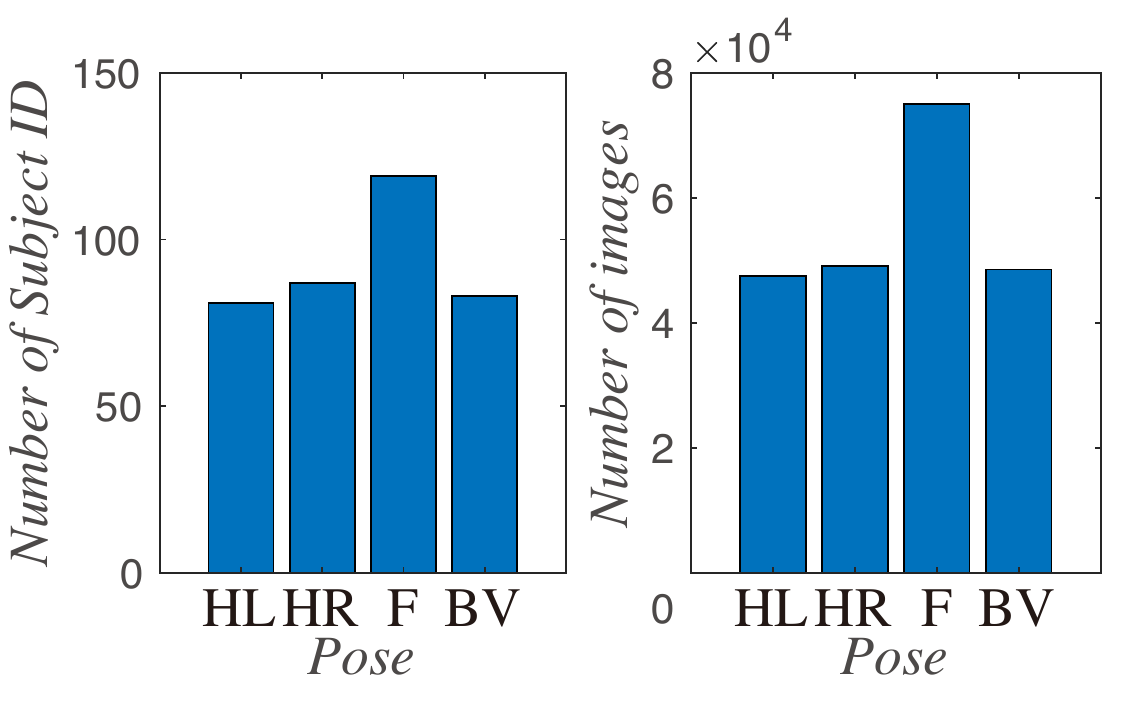}\tabularnewline
\end{tabular}\tabularnewline
(a) & (b) \tabularnewline
\end{tabular}\caption{\label{fig:Cameras-used-to}(a) Cameras used to collect facial
expressions. (b) Distributions of subject ID and images over poses.}
\par\end{centering}
\end{figure}
We create the $\mathrm{F}^{2}$ED dataset in 3 steps as in Fig.~\ref{fig:medb}(a).

\noindent \textbf{Data Collection}. It takes us totally six months
to collect video data. We invite more than 200 different candidates
who are unfamiliar with our research topics. Each candidate is
captured by four cameras placed at four different orientations to
collect videos for persons as shown in Fig.~\ref{fig:Cameras-used-to}
(a). The four orientations are front, half left, half right and bird
view. The half left and half right cameras have a horizontal angle
of 45 degrees with the front of the person, respectively. The bird
view camera has a vertical angle of 30 degrees with the front of the
person. Each camera takes 25 frames per second. The whole video capturing
process is designed as a normal conversation between the candidate
and two psychological experts. Totally, we aim at capturing 54 different
types of expressions \cite{Lee2017Reading}, \emph{e.g.}, acceptance,
angry, bravery, calm, disgust, envy, fear, neutral and so on. The
conversation will follow some scripts which are calibrated by psychologists,
and thus can induce/inspire one particular type of expression
successfully conveyed by the candidates. For each candidate, we only
save 5 minutes' video segment for each type of emotion.

\noindent \textbf{Data Processing}. With gathered expression videos,
we further generate the final image dataset by human review, key images
generation and face alignment. Specifically, the human review step
is very important to guarantee the general quality of recorded expressions.
Three psychologists are invited to help us review the captured emotion
videos. Particularly, each captured video will be labeled by these
psychologists. We only save the videos that have consistent labels by the psychologists.
Thus totally about 119 identities' videos are preserved finally. Then key frames are extracted from each resulting video and face detection and alignment are conducted by the toolboxes of
Dlib and MTCNN \cite{zhang2016joint} over each frame. Critically,
the face bounding boxes are cropped from the original images and resized
to a resolution of $256\times256$ pixels. Finally we get the dataset
$\mathrm{F}^{2}$ED of totally $219719$ images.

\subsection{Statistics and Meta-information of $\mathrm{F}^{2}$ED }

\noindent \textbf{Data} Information. There are 4 types of face information
in our dataset, including person identity, facial expression, pose and landmarks.

\noindent \textbf{Person Identity.} Totally we have 119 persons, including
37 male and 82 female aging from 18 to 24. Most of them are university
students. Each person expresses his/her emotions under guidance and
the video is taken when the person's emotion is observed.

\noindent \textbf{Facial expression.} Our dataset is composed of
54 types of emotions, based on the theory of Lee \cite{Lee2017Reading}.
In this work, it expands the emotion set of Plutchik by including
more complex mental states based on seven eye features. The seven
features include temporal wrinkles, wrinkles below eyes, nasal wrinkles,
brow slope, brow curve, brow distance and eye apertures. The 54
emotions can be clustered into 4 groups by k-means clustering algorithm
as shown in Fig.~\ref{fig:medb}(b). We also compute data distribution
in Fig.~\ref{fig:dist_on_expr}.

\begin{figure*}
\begin{centering}
\begin{tabular}{cc}
\hspace{-0.3in}%
\begin{tabular}{c}
\includegraphics[width=0.4\textwidth]{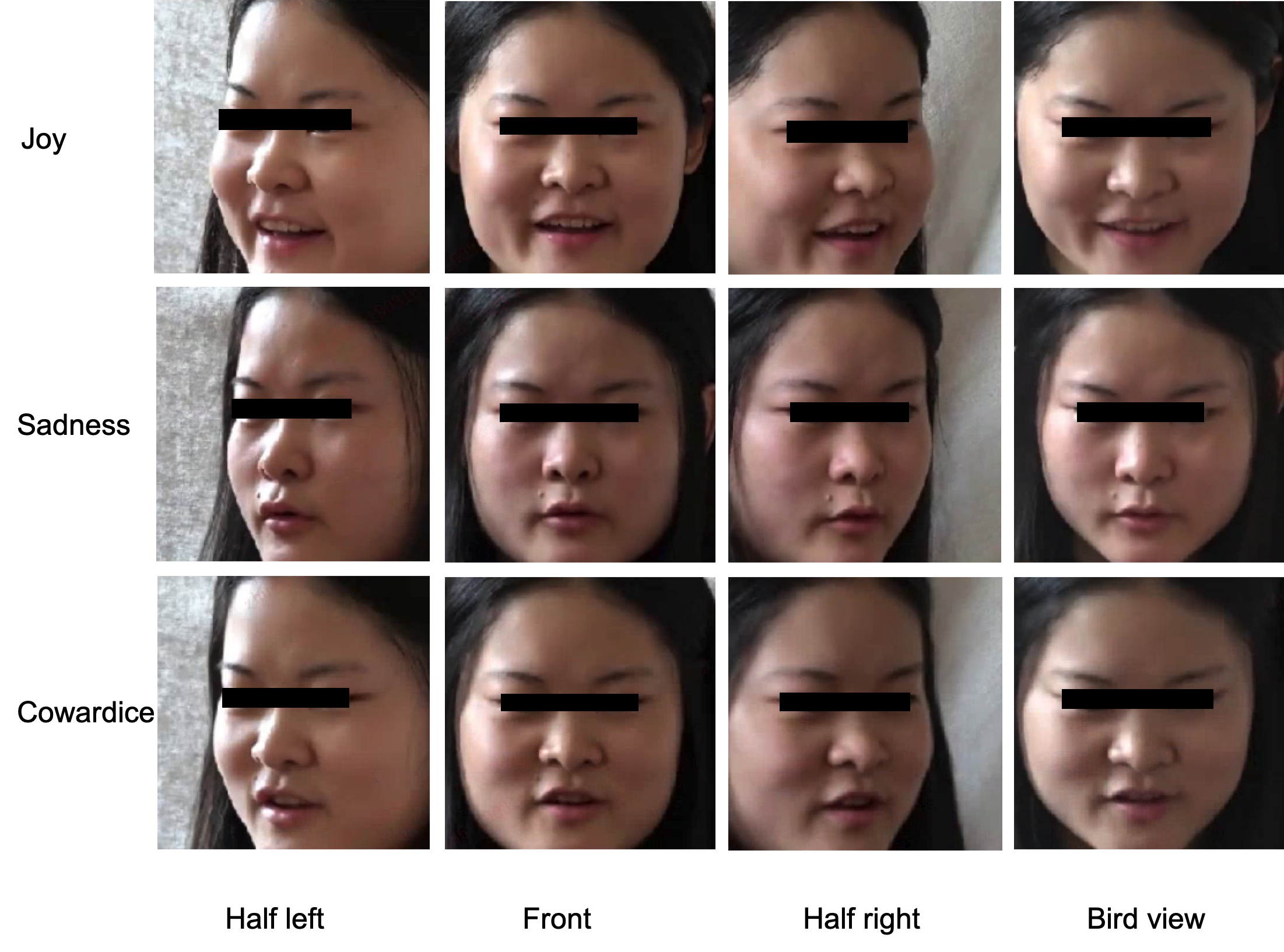}\tabularnewline
\end{tabular} & \hspace{-0.3in}%
\begin{tabular}{c}
\includegraphics[width=0.5\textwidth]{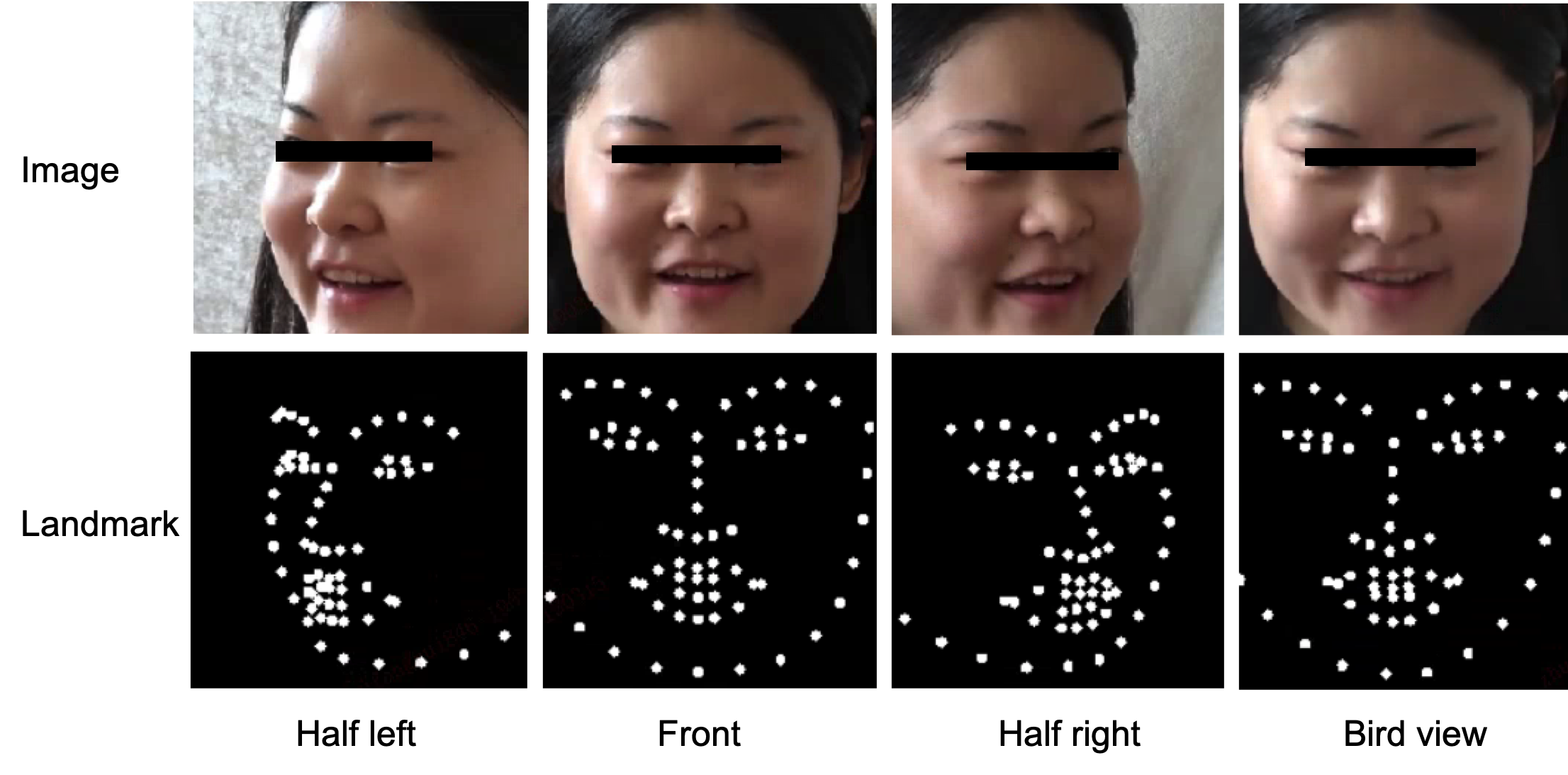}\tabularnewline
\end{tabular}\tabularnewline
(a) Face examples  & (b) Facial landmark examples\tabularnewline
\end{tabular}
\par\end{centering}
\caption{\label{fig:examples_landmark_dist}(a) There are some facial examples
of $\mathrm{F}^{2}$ED with different poses and emotions. (b) We give the facial landmark examples as
the meta-information of $\mathrm{F}^{2}$ED.}

\end{figure*}
\noindent \textbf{Pose. }As an important type of meta-information,
poses often cause facial appearance changes. In real world applications,
facial pose variations are mainly introduced by the relative position
and orientation changes of the cameras to persons. In $\mathrm{F}^{2}$ED,
we collect videos from 4 orientations: half left, front, half right
and bird view. Fig.~\ref{fig:examples_landmark_dist}(a) gives some
examples of the $\mathrm{F}^{2}$ED of different poses. In $\mathrm{F}^{2}$ED
we have 47053 half left, 49152 half right, 74985 front and 48529 bird
view images. The distributions of subject ID and image number over
poses are compared in Fig.~\ref{fig:Cameras-used-to} (b).

\noindent \textbf{Facial Landmarks.} Facial landmarks define the
contour of facial components, including eye, nose, mouth and cheek.
First we extract the facial landmarks with 68 points into position
annotation text files by the Dlib. Then we convert the landmark position
text file into images in a mask style. The example landmark images
are shown in Fig.~\ref{fig:examples_landmark_dist}(b).

\vspace{-0.1in}
Tab.~\ref{tab:Comparison-with-existing} shows the comparison between
our $\mathrm{F}^{2}$ED with existing facial expression database.
As shown in the table, our dataset contains 54 subtle expression types,
while other datasets only contain 7 or 8 expression types. For the
person number, CK+, KDEF and $\mathrm{F}^{2}$ED are nearly the same.
The current public facial expression datasets are usually collected
in two ways: in the wild or in the controlled environment. The FER2013
is collected in the wild, so the number of pose can not be determined.
The rest datasets are collected in a controlled environment, where
the number of pose for CK+ and JAFFE is 1, KDEF is 5 and $\mathrm{F}^{2}$ED
is 4. Our $\mathrm{F}^{2}$ED is the only one that contains the bird
view pose images which is very useful in real world scenario. For
image number, $\mathrm{F}^{2}$ED contains 219719 images, which is
6 times larger than the second largest dataset. All datasets have
a similar resolution except FER2013 which has only a $48\times48$
resolution. CK+ and $\mathrm{F}^{2}$ED are generated from 593 video
sequences and 5418 video sequences.

\section{Learning on $\mathrm{F}^{2}$ED }

\subsection{Learning tasks\label{subsec:Learning-tasks-on}}

In the $\mathrm{F}^{2}$ED, we consider the expression learning
over different types of variants as shown in Fig.~\ref{fig:medb}(c);
and further study the influence of different poses and subjects
over the FER. To the best of our knowledge, this is the first exploration on this type of tasks. Particularly, we are interested in the following tasks for this dataset.

\noindent \textbf{Expression recognition in the standard setting}
(ER-SS). The first and most important task is to directly learn the
supervised classifiers on $\mathrm{F}^{2}$ED. Particularly, as shown
in Fig.~\ref{fig:Cameras-used-to}(b) and Fig.~\ref{fig:dist_on_expr},
our dataset has balanced number of pose and emotion classes. We thus randomly shuffle our dataset and split it into 175000, 19719 and
25000 images for the train, validation and test set, respectively.
The classifiers should be trained and validated on the train and validation
sets, and predicted over the test set.

\noindent \textbf{Expression recognition with unbalanced expression distribution
(ER-UE). }We further compare the results of learning classifiers with
unbalanced facial expressions. In real word scenario, some facial
expressions are rare, \emph{e.g}., cowardice. Thus it is imperative
to investigate the FER in such an unbalanced expression setting. Specifically,
we take 20\% of total facial expressions as the rare classes. Among
these rare classes, 90\% of the images are kept as the testing
instances, the rest 10\% are used as the train set. The other 80\% classes
are treated as the normal emotion classes, and all of them are used for training. Thus, totally we have 178989 and 140730 images for the
train and test set, respectively. For expression type analysis, there
are 54 expression types in train set and 11 expression types in test
set. On average, the occurrence frequency of testing expression class is only $1/10$ of that of training classes. In our setting, we assume that
the model works with the prior knowledge that there are 54 rather than 11 expression
classes in testing, which makes the chance
of ER-UE task keep $1/54$.

\noindent \textbf{Expression recognition with unbalanced poses (ER-UP)}.
The learning task is further conducted with unbalanced poses. In this setting, we assume that the half left pose
is rare in the train set. Thus the 10\% of the half left pose images are used as the train set, and the rest 90\% are used as test set. The other three types of poses
-- the half right, front, bird view pose images are used as the train
set. Thus we get 177372 training images and 42347 testing images.
For pose type analysis, there are 4 poses in train set and 1 pose
in test set. This task aims to predict the expressions with rare poses in training set.

\noindent \textbf{Expression recognition with zero-shot ID (ER-ZID).}
We aim at recognizing the expression types of the persons that have not been seen before. Particularly, we randomly pick the
images from 21 and 98 persons as train and test set, respectively.
This results in 189306 training and 30413 testing images. The task is to recognize expressions with zero-shot ID, referring to the disjoint subject ID in train and test sets. This enables us to verify whether the model can learn the person invariant feature for emotion classification.

\subsection{Learning methods}

\begin{figure*}
\centering{}\includegraphics[width=0.8\textwidth]{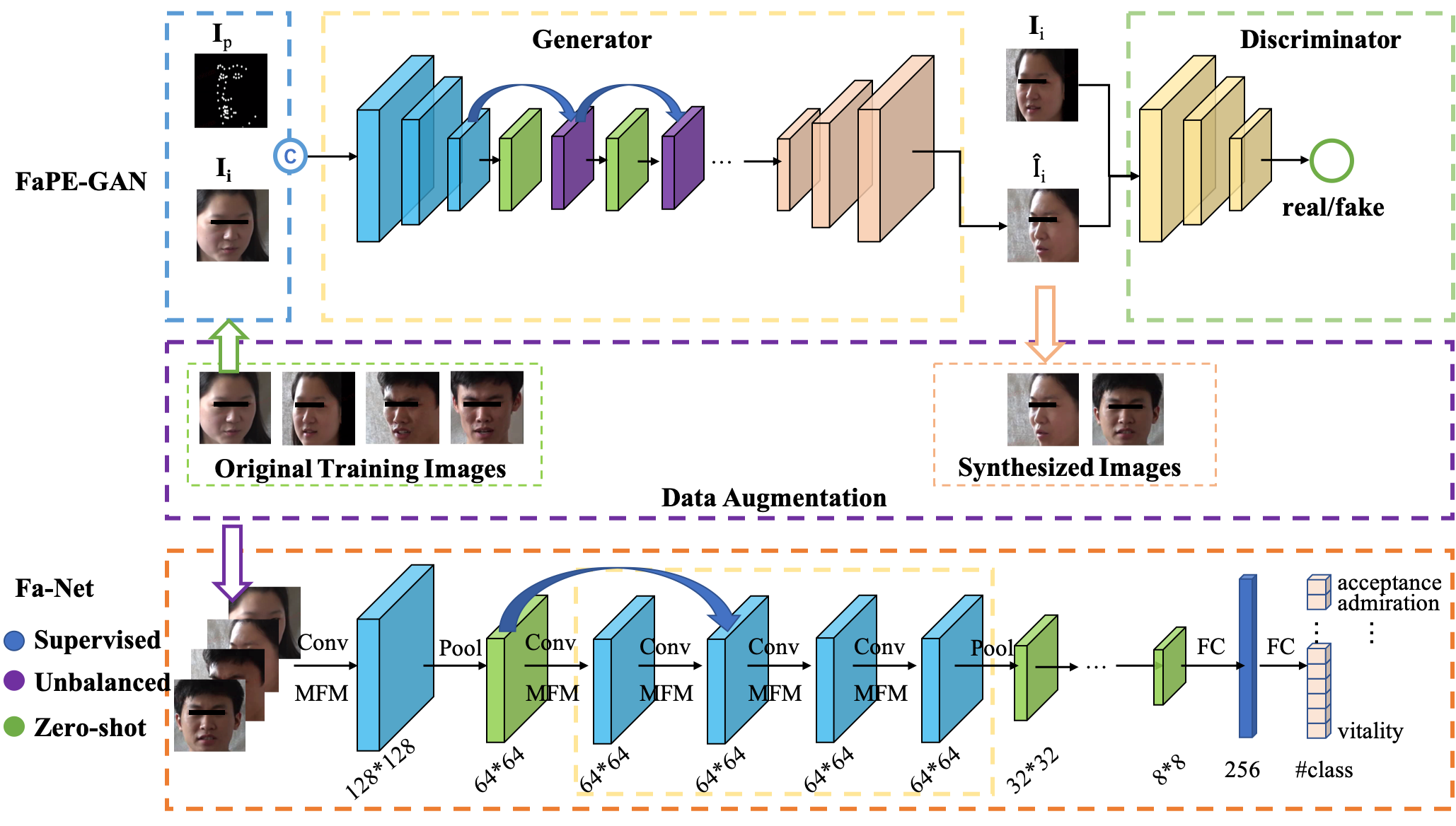}\caption{\label{fig:The-framework-of}Overview of our framework. It includes
the FaPE-GAN and Fa-Net component. FaPE-GAN can synthesize face images
with input image and target pose. The Fa-Net is the classification
network which is trained by the augmented face images and original
face images. The Fa-Net can be applied in supervised, unbalanced and zero-shot learning.}
\end{figure*}

We propose an end-to-end framework to address the four learning tasks
in Fig.~\ref{fig:The-framework-of}. Particularly, to tackle the
issues of learning unbalanced number of images, our key idea is to employ
the GAN based models for data augmentation to produce balanced training set. Our framework
has the components of Facial Pose GAN (FaPE-GAN), and Face classification
Networks (Fa-Net). The former one is an image synthesis network, and the
latter is a classification network.

\begin{figure}
\begin{centering}
\includegraphics[width=0.45\textwidth]{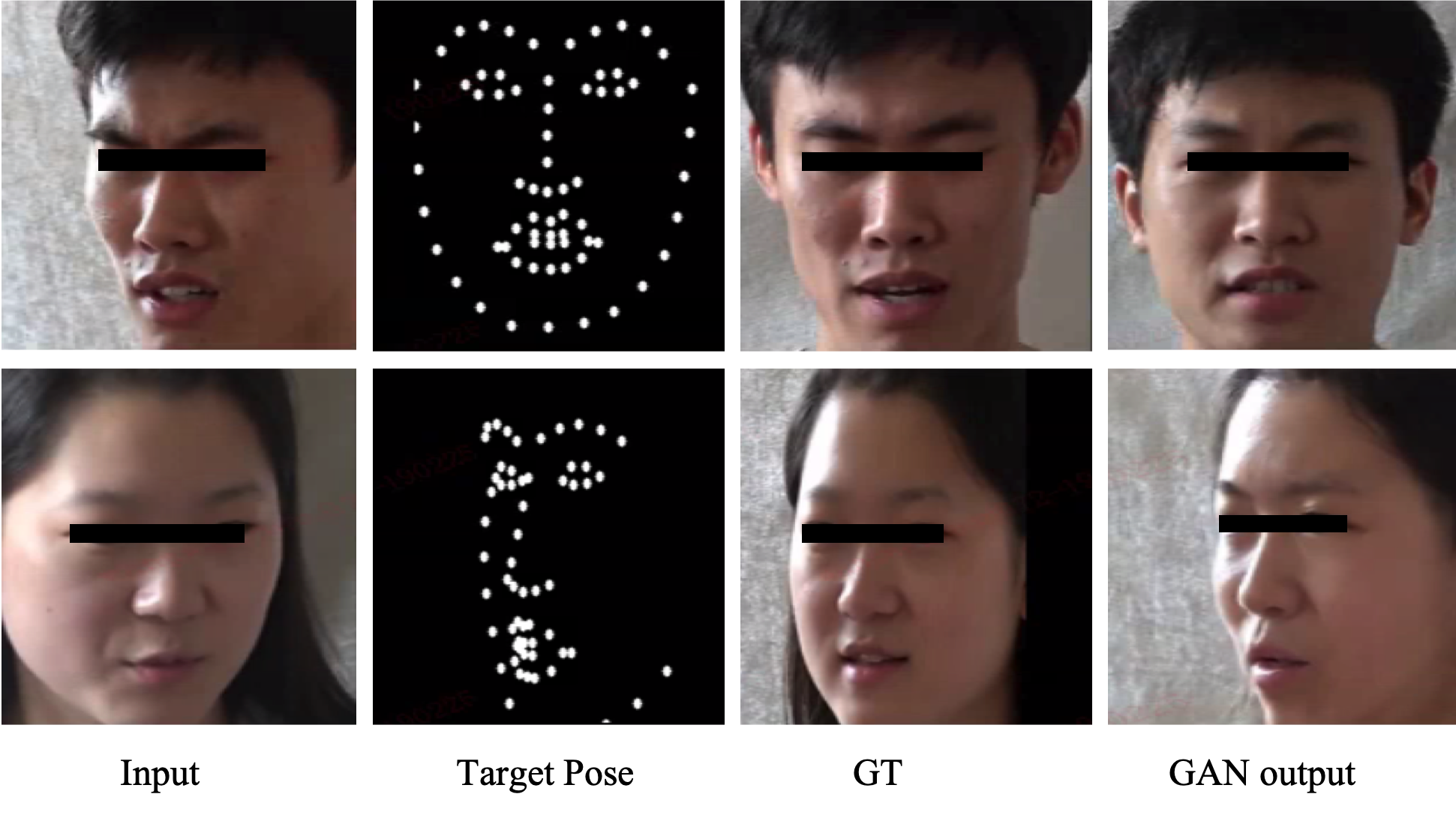}\caption{\label{fig:GAN-output-examples}GAN output examples}
\par\end{centering}
\end{figure}

\noindent \textbf{FaPE-GAN}. It is trained by a combination of the
training images and synthesized face images of new poses. The facial
poses are normally represented by a landmark set. As shown in Fig.~\ref{fig:The-framework-of}, this network firstly takes the face image $\mathbf{I}_{i}$ and the pose image $\mathbf{I}_{\mathcal{P}}$ as input, then the generator produces
the fake image $\hat{\mathbf{I}}_{i}$ of the same person with the
pose of $\mathbf{I}_{\mathcal{P}}$, \emph{i.e}., $\hat{\mathbf{I}}_{i}=G_{FaPE}\left(\mathbf{I}_{i},\mathbf{I}_{\mathcal{P}}\right)$, and the discriminator tries to differentiate the fake target image
$\hat{\mathbf{I}}_{i}$ from the real input image $\mathbf{I}_{i}$.
Despite the pose may be changed in $\hat{\mathbf{I}}_{i}$, our FaPE-GAN
still aims to keep the face identity of $\mathbf{I}_{i}$. Critically,
we introduce the adversarial loss as,

\begin{align}
\underset{G}{\mathrm{min}}\underset{D}{\mathrm{max}}\,\mathcal{L}_{GAN}   & = \mathbb{E}_{\mathbf{I_{i}}\sim p_{d}\left(\mathbf{I}_{i}\right)}\left[\mathrm{log}\,D\left(\mathbf{I_{i}}\right)\right]\\
+ & \left[\mathrm{log}\,\left(1-D\left(G_{FaPE}\left(\mathbf{I_{i},I_{\mathcal{P}}}\right)\right)\right)\right]\label{eq:ad-loss}
\end{align}

\noindent where $p_{d}\left(\mathbf{I}_{i}\right)$ are the distributions
of real images $\mathbf{I}_i$. The training process iteratively updates
the parameters of generator $G_{FaPE}$ and discriminator $D$. The generator loss can be formulated as,

\begin{equation}
\mathcal{L}_{G_{FaPE}}=\mathcal{L}_{GAN}+\lambda\mathcal{L}_{L_{1}}\label{eq:loss_G}
\end{equation}

\noindent where we have $\mathcal{L}_{L_{1}}=\mathbb{E}_{\mathbf{I}_{t}\sim p_{d}\left(\mathbf{I}_{t}\right)}\left[\left|\mathbf{I}_{t}-\hat{\mathbf{I}}_{i}\right|\right]$,  $\mathbf{I}_{t}$ is the real target image
and $\hat{\mathbf{I}}_{i}=G_{Dec}\left(G_{Enc}\left(\mathbf{I}_{i},\mathbf{I}_{\mathcal{P}}\right)\right)$
is the reconstructed image, with the input image $\mathbf{I}_{i}$
and facial pose $\mathbf{I}_{\mathcal{P}}$. \cite{mirza2014conditional}.
The hyperparameter $\lambda$ is used to balance the two terms. The discriminator loss is formulated as, $\mathcal{L}_{D}=-\mathcal{L}_{GAN}$.
The training process iteratively optimizes the loss functions of $\mathcal{L}_{G_{FaPE}}$
and $\mathcal{L}_{D}$. Fig.~\ref{fig:GAN-output-examples} shows
two examples generated by FaPE-GAN .

\noindent \textbf{Fa-Net}. The same classification network is utilized to address all the four learning tasks in Sec.~\ref{subsec:Learning-tasks-on}.
Particularly, the backbone network is LightCNN \cite{Xiang2015A}. The $G_{FaPE}$ can synthesize plenty of additional face images in
alleviating the issues of unbalanced training images. The augmented faces and original input faces are thus used to train our classification
network. \textcolor{red}{}

\section{Experiments}

Extensive experiments are conducted on $\mathrm{F}^{2}$ED to evaluate
the learning tasks defined in Sec.~\ref{subsec:Learning-tasks-on}.
Furthermore, the tasks of facial emotion recognition are also evaluated on FER2013 and JAFFE dataset.

\noindent \textbf{Implementation details.} The $\lambda$ is set
to 10, and the Adam optimizer is used in learning the FaPE-GAN with
the learning rate of $2e-4$. The $\beta_{1}$ and $\beta_{2}$ are
set as 0.5 and 0.999 respectively. The training epoch number is set
to 100. For the facial expression classification network, We use the
SGD optimizer with a momentum of 0.9 and decrease the learning rate
by 0.457 every 10 steps. The max epoch number is set to 100. The learning
rate and batch size varies depending on the dataset size. To train
the classification model, we set the learning rate/batch size as 0.01/128,
2$e-3$/64 and $5e-4$/32, on $\mathrm{F}^{2}$ED, FER2013 and JAFFE, respectively.

\subsection{Results on FER2013 dataset}

\noindent \textbf{Settings}. Following the setting of ER-SS, we conduct
the experiments on FER2013 by using the entire 28709 training images
and 3589 validation images to train/validate our model, which is further
tested on the rest 3589 test images. The FER classification accuracy
is reported as the evaluation metric to compare different competitors.

\noindent \textbf{Competitors}. Our model is compared against several
competitors, including Bag of Words~\cite{ionescu2013local}, VGG+SVM~\cite{georgescu2018local},
GoogleNet~\cite{giannopoulos2018deep}, Mollahosseini \emph{et al~}\cite{mollahosseini2016going},  DNNRL~\cite{guo2016deep}
and Attention CNN~\cite{minaee2019deep}. Classifiers based on hand-crafted
features, or specially designed architectures for FER, are investigated
here. These methods can achieve the state-of-the-art results on this
dataset.

\begin{figure}
\centering{}%
\begin{tabular}{cc}
\hspace{-0.3in}\includegraphics[width=0.27\textwidth]{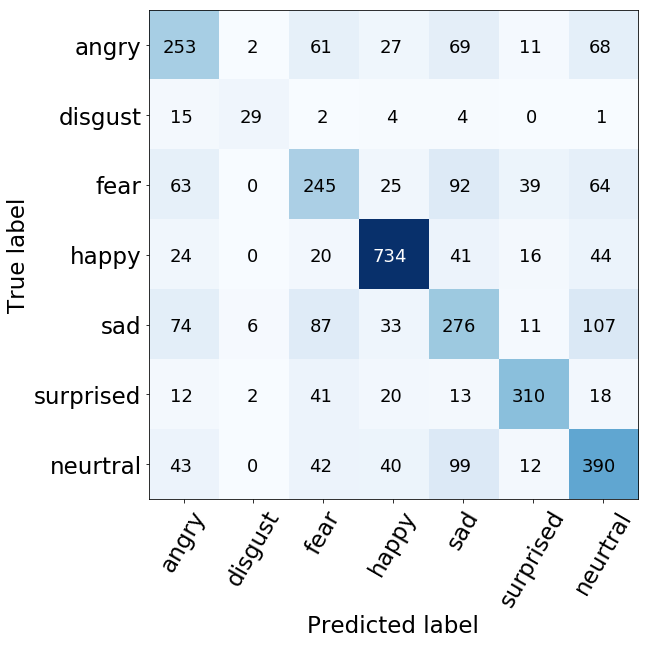} & \hspace{-0.2in}\includegraphics[width=0.27\textwidth]{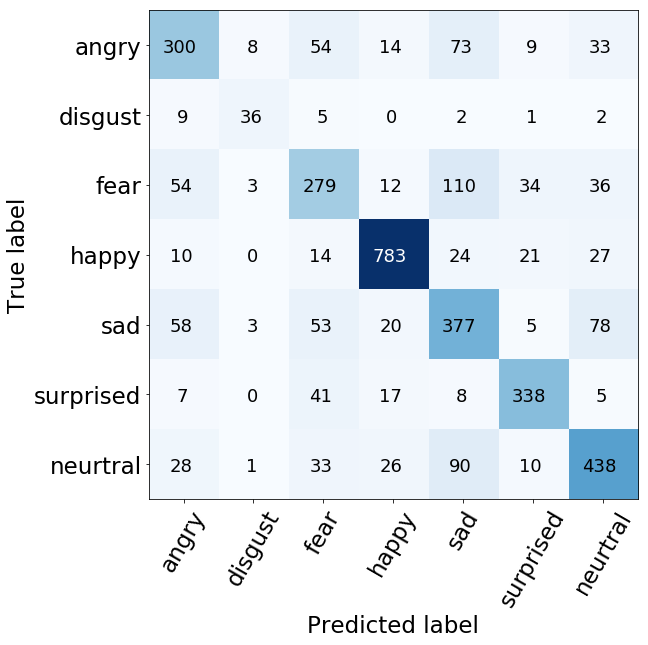}\tabularnewline
(a)  & (b) \tabularnewline
\end{tabular}\caption{\label{fig:fer2013_sl_cm}(a) The confusion matrix on FER 2013 for
Fa-Net without pre-training. (b) The confusion matrix on FER2013 for
Fa-Net pre-trained on $\mathrm{F}^{2}$ED}
\end{figure}

\noindent \textbf{Results on FER2013}. To show the efficacy of our dataset, our classification network -- Fa-Net is pre-trained on our
$\mathrm{F}^{2}$ED, and then fine-tuned on the training set of FER2013 dataset. The results show that our model can achieve the accuracy of 71.1\%,
which is superior to other state-of-the-art methods, as compared in
Tab.~\ref{tab:Acc_FER2013_SL}. Tab.~\ref{tab:pretrain comparison}
shows that the Fa-Net pre-trained on $\mathrm{F}^{2}$ED can improve the
expression recognition performance by 8.8\% comparing to the one
without pre-training. The confusion matrix in Fig.~\ref{fig:fer2013_sl_cm}
shows that pre-training increases the scores on all expression types.
It demonstrates that the $\mathrm{F}^{2}$ED dataset with large expression
variations from more persons can pre-train a deep network with good initialization parameters. Note that our Fa-Net is not specially designed for FER task, since our Fa-Net is built upon the backbone -- LightCNN, one typical face recognition architecture.

\begin{table}
\centering{}%
\begin{tabular}{c|c}
\hline
Model & Acc.\tabularnewline
\hline
\hline
Bag of Words~\cite{ionescu2013local} & 67.4\%\tabularnewline
\hline
VGG+SVM~\cite{georgescu2018local} & 66.3\%\tabularnewline
\hline
GoogleNet~\cite{giannopoulos2018deep} & 65.2\%\tabularnewline
\hline
Mollahosseini \emph{et al~}\cite{mollahosseini2016going} & 66.4\%\tabularnewline
\hline
DNNRL~\cite{guo2016deep} & 70.6\%\tabularnewline
\hline
Attention CNN~\cite{minaee2019deep} & 70.0\%\tabularnewline
\hline
\hline
Fa-Net & 71.1\%\tabularnewline
\hline
\end{tabular}\caption{\label{tab:Acc_FER2013_SL}Accuracy on FER2013 test set in supervised
learning setting}
\end{table}

\subsection{Results on JAFFE dataset}

\noindent \textbf{Settings}. For the setting of ER-SS, we follow the split setting of the deep-emotion
paper\cite{minaee2019deep} to use 120 images for training, 23 images
for validation, and keep totally 70 images for test (7 emotions per face ID).

\noindent \textbf{Competitors}. Our model is compared against several
competitors, including Fisherface\cite{abidin2012neural}, Salient
Facial Patch \cite{happy2015automatic}, CNN+SVM\cite{shima2018image}
and Attention CNN~\cite{minaee2019deep}.  These methods are tailored
for the tasks of FER.

\begin{figure}
\centering{}%
\begin{tabular}{cc}
\hspace{-0.3in}\includegraphics[width=0.26\textwidth]{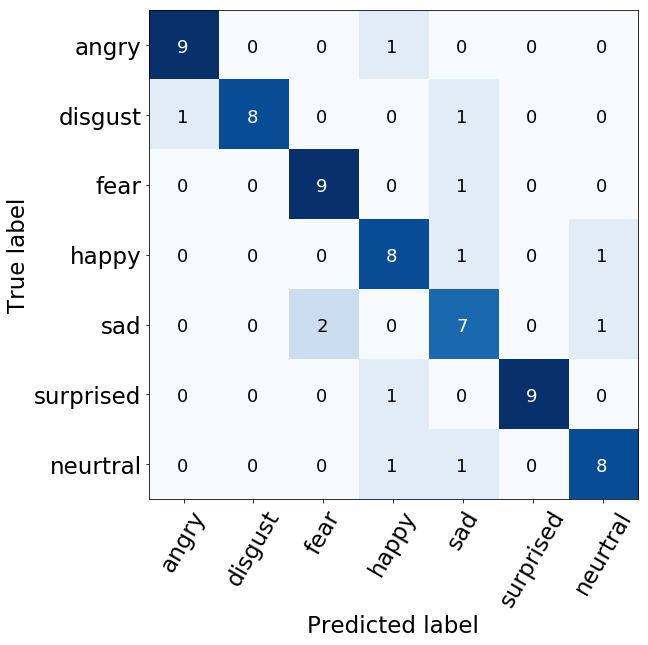} & \hspace{-0.2in}\includegraphics[width=0.26\textwidth]{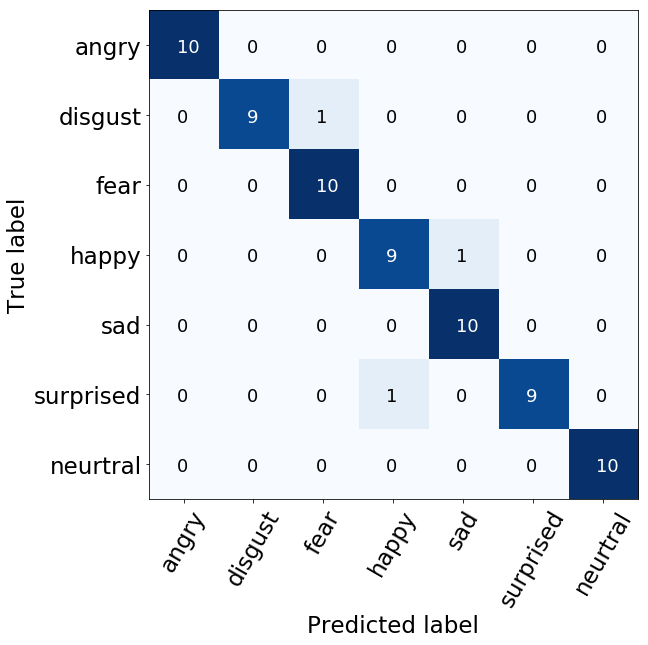}\tabularnewline
(a)  & (b) \tabularnewline
\end{tabular}\caption{\label{fig:jaffe_sl_cm}(a) The confusion matrix on FER 2013 for Fa-Net
without pre-training. (b) The confusion matrix on FER2013 for Fa-Net pre-trained on $\mathrm{F}^{2}$ED}
\end{figure}

\noindent As listed in Tab.~\ref{tab:Acc_JAFFE_SL}, our model achieved
the accuracy of 95.7\%, outperforming all the other competitors.
Remarkably, our model surpasses the Attention CNN by 2.9\% in the
same data split setting. The accuracy of CNN+SVM is slightly lower than our model by 0.4\%, even though their model is trained and tested on the entire dataset. This shows the efficacy of our dataset
in pre-training the network. Tab.~\ref{tab:pretrain comparison}
further shows that Fa-Net pre-trained on the $\mathrm{F}^{2}$ED has
clearly improved the performance by 12.8\%. The confusion matrix in
Fig.~\ref{fig:jaffe_sl_cm} shows that the pre-trained Fa-Net only
makes 3 wrong predictions and surpasses the one without pre-training on all expression types.

\begin{table}
\centering{}%
\begin{tabular}{c|c}
\hline
Model & Acc.\tabularnewline
\hline
\hline
Fisherface\cite{abidin2012neural} & 89.2\%\tabularnewline
\hline
Salient Facial Patch\cite{happy2015automatic} & 92.6\%\tabularnewline
\hline
CNN+SVM\cite{shima2018image} & 95.3\%\tabularnewline
\hline
Attention CNN\cite{minaee2019deep} & 92.8\%\tabularnewline
\hline
\hline
Fa-Net & 95.7\%\tabularnewline
\hline
\end{tabular}\caption{\label{tab:Acc_JAFFE_SL}Accuracy on JAFFE test set in supervised
learning setting.}
\end{table}

\subsection{Results on $\mathrm{F}^{2}$ED }

\textbf{Results on our dataset}. We conduct the four different learning
tasks on our dataset, namely, supervised (ER-SS), unbalanced expression
(ER-UE), unbalanced pose (ER-UP) and zero-shot ID (ER-ZID) by the
data split setting described in Sec.~\ref{subsec:Learning-tasks-on}.
Note that, since our Fa-Net is built upon the general face recognition
backbone -- LightCNN, it can thus be served as the main network in
our experiments.

\noindent \textbf{ER-SS task}. Our model has achieved the accuracy
of 73.6\% as shown in Tab.~\ref{tab:Acc_ours}. It shows that our
$\mathrm{F}^{2}$ED is well annotated so it can be used for classification
task. Considering the large scale of the facial expression dataset,
this performance is already very good and difficult to obtain which
demonstrates that LightCNN is a good backbone for facial recognition.
By using FaPE-GAN for data augmentation, the performance of our model
is further improved by 0.9\% comparing to the Fa-Net without GAN,
which means that GAN is useful to generate more diversified examples
for training.

\begin{table}
\begin{centering}
\begin{tabular}{c|c|c}
\hline
Dataset & Pre-trained & Acc.\tabularnewline
\hline
\hline
\multirow{2}{*}{FER2013} &  & 62.3\%\tabularnewline
\cline{2-3}
 & $\checkmark$ & 71.1\%\tabularnewline
\hline
\hline
\multirow{2}{*}{JAFFE} &  & 82.9\%\tabularnewline
\cline{2-3}
 & $\checkmark$ & 95.7\%\tabularnewline
\hline
\end{tabular}
\par\end{centering}
\caption{\label{tab:pretrain comparison}Results of the Fa-Net model with and without
pre-trained on our $\mathrm{F}^{2}$ED. }
\end{table}

\noindent \textbf{ER-UE task}. The accuracy of direct classification
is 30.8\% as shown in Tab.~\ref{tab:Acc_ours}. This shows that the
propose ER-UE task is very difficult, as the FER task greatly
suffers from the unbalanced emotion data. Particularly, in our setting,
only 10\% examples from the 11 facial expression types appear in the
training set, and the classifiers are thus confused by the other
43 emotion classes in the training stage. Furthermore, we also show
that the data augmentation strategy endowed by our FaPE-GAN can indeed
help to improve the performance of FER: the performance is improved by
3.5\% which is larger than the 0.9\% improvement in supervised learning
setting. This indicates that the data augmentation is more effective
in the data sparse condition such as unbalanced learning.

\noindent \textbf{ER-UP task}. Towards this task, our Fa-Net can
hit the accuracy of 39.9\% as shown in Tab.~\ref{tab:Acc_ours}.
Again, we argue that the proposed ER-UP is a very hard task, since
this accuracy is only slightly better than the performance of ER-UE.
This shows that the unbalanced pose data may also negatively affect the
performance of FER task.
Essentially, there are 54 types of expressions which are more diversified
than the pose. Our data augmentation can still work in such a setting,
and the synthesized data can help to train the Fa-Net, and alleviate
the problem of unbalanced poses. As a result, it improves the performance
of Fa-Net by 3.6\%.

\noindent \textbf{ER-ZID task}. Surprisingly, the learning task proposed
in this setting is the most challenging one compared with the other
learning tasks. As shown in Tab.~\ref{tab:Acc_ours}, we notice that
our model only achieves an accuracy of 7.1\% while the chance in fact is 1.9\% ($\frac{1}{54}$ as described before), since the zero-shot
task is much more difficult than the unbalanced task. This indicates
that the generalization ability of FER is subject to other persons
that the model has never seen before. Actually, this is the most desirable property of the FER model, since one can not assume
the faces of test persons always appear in the training set. In our ER-ZID
task, only 21 persons in the test set are never seen in the training set.
Interestingly, our FaPE-GAN based data augmentation still contributes
a 0.4\% performance improvement over the baseline. This suggests the
data augmentation may be still a potential useful strategy to facilitate the training of classification network.

Overall, our classification model with FaPE-GAN based data augmentation
has clearly surpasses the one without FaPE-GAN on all 4 task types.

\begin{table}
\centering{}\hspace{-0.3in}%
\begin{tabular}{c||c|c|c|c}
\hline
model/acc & ER-SS & ER-UE & ER-UP & ER-ZID\tabularnewline
\hline
\hline
Fa-Net & 72.7 & 27.3 & 36.3 & 6.7\tabularnewline
\hline
FaPE-GAN+Fa-Net & 73.6 & 30.8 & 39.9 & 7.1\tabularnewline
\hline
\end{tabular}\caption{\label{tab:Acc_ours}Accuracy on $\mathrm{F}^{2}$ED for Fa-Net with and without data augmentation in supervised(ER-SS),
unbalanced expression(ER-UE), unbalanced pose(ER-UP) and zero-shot ID(ER-ZID) setting}
\end{table}

\section{Conclusion}

In this work, we introduce $\mathrm{F}^{2}$ED, a new facial expression
database containing 54 different emotion types and more than
200k examples. Furthermore we propose an end-to-end deep neural network
based facial expression recognition framework, which uses a facial pose generative adversarial network to augment the data set. We perform supervised, zero-shot and unbalanced learning tasks
on our $\mathrm{F}^{2}$ED dataset, and the results show that our model
has achieved the state-of-the-art. Subsequently, we fine-tune our
model pre-trained on $\mathrm{F}^{2}$ED on the existing FER2013 and JAFFE
database, and the results demonstrate the efficacy of our $\mathrm{F}^{2}$ED dataset.

{\small{}\bibliographystyle{ieee}
\bibliography{egbib}

\begin{thebibliography}{10}\itemsep=-1pt

\bibitem{abidin2012neural}
Z.~Abidin and A.~Harjoko.
\newblock A neural network based facial expression recognition using
  fisherface.
\newblock {\em International Journal of Computer Applications}, 59(3), 2012.

\bibitem{aneja2016modeling}
D.~Aneja, A.~Colburn, G.~Faigin, L.~Shapiro, and B.~Mones.
\newblock Modeling stylized character expressions via deep learning.
\newblock In {\em Asian Conference on Computer Vision}, pages 136--153.
  Springer, 2016.

\bibitem{bartlett2005recognizing}
M.~S. Bartlett, G.~Littlewort, M.~Frank, C.~Lainscsek, I.~Fasel, and
  J.~Movellan.
\newblock Recognizing facial expression: machine learning and application to
  spontaneous behavior.
\newblock In {\em 2005 IEEE Computer Society Conference on Computer Vision and
  Pattern Recognition (CVPR'05)}, volume~2, pages 568--573. IEEE, 2005.

\bibitem{berretti20113d}
S.~Berretti, B.~B. Amor, M.~Daoudi, and A.~Del~Bimbo.
\newblock 3d facial expression recognition using sift descriptors of
  automatically detected keypoints.
\newblock {\em The Visual Computer}, 27(11):1021, 2011.

\bibitem{corneanu2016survey}
C.~A. Corneanu, M.~O. Sim{\'o}n, J.~F. Cohn, and S.~E. Guerrero.
\newblock Survey on rgb, 3d, thermal, and multimodal approaches for facial
  expression recognition: History, trends, and affect-related applications.
\newblock {\em IEEE transactions on pattern analysis and machine intelligence},
  38(8):1548--1568, 2016.

\bibitem{ekman1997face}
R.~Ekman.
\newblock {\em What the face reveals: Basic and applied studies of spontaneous
  expression using the Facial Action Coding System (FACS)}.
\newblock Oxford University Press, USA, 1997.

\bibitem{georgescu2018local}
M.-I. Georgescu, R.~T. Ionescu, and M.~Popescu.
\newblock Local learning with deep and handcrafted features for facial
  expression recognition.
\newblock {\em arXiv preprint arXiv:1804.10892}, 2018.

\bibitem{giannopoulos2018deep}
P.~Giannopoulos, I.~Perikos, and I.~Hatzilygeroudis.
\newblock Deep learning approaches for facial emotion recognition: A case study
  on fer-2013.
\newblock In {\em Advances in Hybridization of Intelligent Methods}, pages
  1--16. Springer, 2018.

\bibitem{goodfellow2014generative}
I.~Goodfellow, J.~Pouget-Abadie, M.~Mirza, B.~Xu, D.~Warde-Farley, S.~Ozair,
  A.~Courville, and Y.~Bengio.
\newblock Generative adversarial nets.
\newblock In {\em Advances in neural information processing systems}, pages
  2672--2680, 2014.

\bibitem{guo2016deep}
Y.~Guo, D.~Tao, J.~Yu, H.~Xiong, Y.~Li, and D.~Tao.
\newblock Deep neural networks with relativity learning for facial expression
  recognition.
\newblock In {\em 2016 IEEE International Conference on Multimedia \& Expo
  Workshops (ICMEW)}, pages 1--6. IEEE, 2016.

\bibitem{happy2015automatic}
S.~Happy and A.~Routray.
\newblock Automatic facial expression recognition using features of salient
  facial patches.
\newblock {\em IEEE transactions on Affective Computing}, 6(1):1--12, 2015.

\bibitem{inbalanced_data}
C.~Huang, Y.~Li, C.~C. Loy, and X.~Tang.
\newblock Learning deep representation for imbalanced classification.
\newblock In {\em CVPR}, 2016.

\bibitem{ionescu2013local}
R.~T. Ionescu, M.~Popescu, and C.~Grozea.
\newblock Local learning to improve bag of visual words model for facial
  expression recognition.
\newblock In {\em Workshop on challenges in representation learning, ICML},
  2013.

\bibitem{kanade2000comprehensive}
T.~Kanade, Y.~Tian, and J.~F. Cohn.
\newblock Comprehensive database for facial expression analysis.
\newblock In {\em fg}, page~46. IEEE, 2000.

\bibitem{khorrami2015deep}
P.~Khorrami, T.~Paine, and T.~Huang.
\newblock Do deep neural networks learn facial action units when doing
  expression recognition?
\newblock In {\em Proceedings of the IEEE International Conference on Computer
  Vision Workshops}, pages 19--27, 2015.

\bibitem{lampert2014attribute}
C.~H. Lampert, H.~Nickisch, and S.~Harmeling.
\newblock Attribute-based classification for zero-shot visual object
  categorization.
\newblock {\em IEEE Transactions on Pattern Analysis and Machine Intelligence},
  36(3):453--465, 2014.

\bibitem{Lee2017Reading}
D.~H. Lee and A.~K. Anderson.
\newblock Reading what the mind thinks from how the eye sees.
\newblock {\em Psychological Science}, 28(4):494, 2017.

\bibitem{liu2017hydraplus}
X.~Liu, H.~Zhao, M.~Tian, L.~Sheng, J.~Shao, S.~Yi, J.~Yan, and X.~Wang.
\newblock Hydraplus-net: Attentive deep features for pedestrian analysis.
\newblock {\em ICCV}, 2017.

\bibitem{liu2015deep}
Z.~Liu, P.~Luo, X.~Wang, and X.~Tang.
\newblock Deep learning face attributes in the wild.
\newblock In {\em Proceedings of the IEEE International Conference on Computer
  Vision}, pages 3730--3738, 2015.

\bibitem{lucey2010extended}
P.~Lucey, J.~F. Cohn, T.~Kanade, J.~Saragih, Z.~Ambadar, and I.~Matthews.
\newblock The extended cohn-kanade dataset (ck+): A complete dataset for action
  unit and emotion-specified expression.
\newblock In {\em 2010 IEEE Computer Society Conference on Computer Vision and
  Pattern Recognition-Workshops}, pages 94--101. IEEE, 2010.

\bibitem{lundqvist1998karolinska}
D.~Lundqvist, A.~Flykt, and A.~{\"O}hman.
\newblock The karolinska directed emotional faces (kdef).
\newblock {\em CD ROM from Department of Clinical Neuroscience, Psychology
  section, Karolinska Institutet}, 91:630, 1998.

\bibitem{lyons1998coding}
M.~Lyons, S.~Akamatsu, M.~Kamachi, and J.~Gyoba.
\newblock Coding facial expressions with gabor wavelets.
\newblock In {\em Proceedings Third IEEE international conference on automatic
  face and gesture recognition}, pages 200--205. IEEE, 1998.

\bibitem{minaee2019deep}
S.~Minaee and A.~Abdolrashidi.
\newblock Deep-emotion: Facial expression recognition using attentional
  convolutional network.
\newblock {\em arXiv preprint arXiv:1902.01019}, 2019.

\bibitem{mirza2014conditional}
M.~Mirza and S.~Osindero.
\newblock Conditional generative adversarial nets.
\newblock {\em arXiv: Learning}, 2014.

\bibitem{mollahosseini2016going}
A.~Mollahosseini, D.~Chan, and M.~H. Mahoor.
\newblock Going deeper in facial expression recognition using deep neural
  networks.
\newblock In {\em 2016 IEEE winter conference on applications of computer
  vision (WACV)}, pages 1--10. IEEE, 2016.

\bibitem{fer2013}
C.~Pierre-Luc and C.~Aaron.
\newblock Challenges in representation learning: Facial expression recognition
  challenge, 2013.

\bibitem{qian2017multi}
X.~Qian, Y.~Fu, Y.-G. Jiang, T.~Xiang, and X.~Xue.
\newblock Multi-scale deep learning architectures for person re-identification.
\newblock In {\em Proceedings of the IEEE International Conference on Computer
  Vision}, pages 5399--5408, 2017.

\bibitem{qian2018pose}
X.~Qian, Y.~Fu, T.~Xiang, W.~Wang, J.~Qiu, Y.~Wu, Y.-G. Jiang, and X.~Xue.
\newblock Pose-normalized image generation for person re-identification.
\newblock In {\em Proceedings of the European Conference on Computer Vision
  (ECCV)}, pages 650--667, 2018.

\bibitem{shan2009facial}
C.~Shan, S.~Gong, and P.~W. McOwan.
\newblock Facial expression recognition based on local binary patterns: A
  comprehensive study.
\newblock {\em Image and vision Computing}, 27(6):803--816, 2009.

\bibitem{shima2018image}
Y.~Shima and Y.~Omori.
\newblock Image augmentation for classifying facial expression images by using
  deep neural network pre-trained with object image database.
\newblock In {\em Proceedings of the 3rd International Conference on Robotics,
  Control and Automation}, pages 140--146. ACM, 2018.

\bibitem{wang2017multi}
Z.~Wang, K.~He, Y.~Fu, R.~Feng, Y.-G. Jiang, and X.~Xue.
\newblock Multi-task deep neural network for joint face recognition and facial
  attribute prediction.
\newblock In {\em Proceedings of the 2017 ACM on International Conference on
  Multimedia Retrieval}, pages 365--374. ACM, 2017.

\bibitem{Xiang2015A}
W.~Xiang, H.~Ran, Z.~Sun, and T.~Tan.
\newblock A light cnn for deep face representation with noisy labels.
\newblock {\em IEEE Transactions on Information Forensics Security},
  PP(99):1--1, 2015.

\bibitem{xu2018heterogeneous}
B.~Xu, Y.~Fu, Y.-G. Jiang, B.~Li, and L.~Sigal.
\newblock Heterogeneous knowledge transfer in video emotion recognition,
  attribution and summarization.
\newblock {\em IEEE Transactions on Affective Computing}, 9(2):255--270, 2018.

\bibitem{yang2018facial}
H.~Yang, U.~Ciftci, and L.~Yin.
\newblock Facial expression recognition by de-expression residue learning.
\newblock In {\em Proceedings of the IEEE Conference on Computer Vision and
  Pattern Recognition}, pages 2168--2177, 2018.

\bibitem{zhang2018joint}
F.~Zhang, T.~Zhang, Q.~Mao, and C.~Xu.
\newblock Joint pose and expression modeling for facial expression recognition.
\newblock In {\em Proceedings of the IEEE Conference on Computer Vision and
  Pattern Recognition}, pages 3359--3368, 2018.

\bibitem{zhang2016joint}
K.~Zhang, Z.~Zhang, Z.~Li, and Y.~Qiao.
\newblock Joint face detection and alignment using multitask cascaded
  convolutional networks.
\newblock {\em IEEE Signal Processing Letters}, 23(10):1499--1503, 2016.

\end{thebibliography}
 }{\small\par}
\end{document}